\documentclass[conference]{IEEEtran}
\IEEEoverridecommandlockouts
\usepackage{amsmath,amssymb,amsfonts}
\usepackage{algorithmic}
\usepackage{graphicx}
\usepackage{textcomp}
\usepackage{csquotes}
\usepackage{xcolor,soul}
\usepackage{adjustbox}
\usepackage{hyperref}
\usepackage{multirow}
\usepackage{titlesec}
\usepackage{subcaption}
\usepackage{makecell}
\usepackage{orcidlink}
\usepackage{placeins}

\usepackage[numbers,sort&compress]{natbib}

\usepackage{etoolbox}

\definecolor{columbiablue}{rgb}{0.61, 0.87, 1.0}
\titlespacing{\subsubsection}{0pt}{0pt}{1pt}

\usepackage{amsmath}

\DeclareMathOperator*{\argmin}{arg\,min}
\usepackage{mathtools}
\usepackage[colorinlistoftodos,prependcaption,textsize=tiny]{todonotes}
\usepackage{gensymb}
\usepackage{multirow}

\usepackage{parskip}

\begin{document}

\title{ResAlignNet: A Data-Driven Approach for INS/DVL Alignment}
\author{\IEEEauthorblockN{Guy Damari\orcidlink{ 
0009-0001-6394-6026} and Itzik Klein\orcidlink{0000-0001-7846-0654}}
\IEEEauthorblockA{{The Hatter Department of Marine Technologies} \\
{Charney School of Marine Sciences, University of Haifa}\\
Haifa, Israel}
}

\maketitle
\begin{abstract}
Autonomous underwater vehicles rely on precise navigation systems that combine the inertial navigation system and the Doppler velocity log for successful missions in challenging environments where satellite navigation is unavailable. The effectiveness of this integration critically depends on accurate alignment between the sensor reference frames. Standard model-based alignment methods between these sensor systems suffer from lengthy convergence times, dependence on prescribed motion patterns, and reliance on external aiding sensors, significantly limiting operational flexibility. To address these limitations, this paper presents ResAlignNet, a data-driven approach using the 1D ResNet-18 architecture that transforms the alignment problem into deep neural network optimization, operating as an in-situ solution that requires only sensors on board without external positioning aids or complex vehicle maneuvers, while achieving rapid convergence in seconds. Additionally, the approach demonstrates the learning capabilities of Sim2Real transfer, enabling training in synthetic data while deploying in operational sensor measurements. Experimental validation using the Snapir autonomous underwater vehicle demonstrates that ResAlignNet achieves alignment accuracy within 0.8° using only 25 seconds of data collection, representing a 65\% reduction in convergence time compared to standard velocity-based methods. The trajectory-independent solution eliminates motion pattern requirements and enables immediate vehicle deployment without lengthy pre-mission procedures, advancing underwater navigation capabilities through robust sensor-agnostic alignment that scales across different operational scenarios and sensor specifications.
\end{abstract}

\section{Introduction}
Underwater navigation systems are critical for a wide range of marine applications, particularly autonomous underwater vehicles (AUVs) operating in challenging environments where global navigation satellite systems (GNSSs) are unavailable \cite{kinsey2006survey}. Integration of inertial navigation systems (INS) and Doppler velocity logs (DVL) has emerged as one of the most promising solutions for precision underwater navigation, without reliance on external infrastructure \cite{stutters2008navigation}. \\
\noindent
INS provides autonomous navigation capability by computing vehicle position, velocity and orientation from gyroscope and accelerometer measurements \cite{groves2013principles}. While providing high-frequency updates and excellent short-term precision, inertial systems suffer from cumulative drift errors, making them unsuitable as standalone navigation solutions for extended operations \cite{titterton2004strapdown, akeila2013reducing}. \\
\noindent
DVLs complement inertial systems by providing drift-free external velocity measurements. The DVL employs a configuration of four acoustic transducers, each emitting a distinct beam toward the seafloor. By analyzing reflected acoustic signals from the seafloor, the DVL provides precise measurements of vehicle velocity \cite{brokloff1994matrix,farrell2008aided}. 
\noindent
However, the effectiveness of INS/DVL integrated navigation systems is critically dependent on the DVL calibration \cite{yampolsky2025dcnet} and the alignment between the inertial measurement unit (IMU) and the DVL reference frames \cite{kinsey2002towards, troni2010new,whitcomb1999advances}.
The alignment problem for underwater navigation arises because, due to mechanical and other constraints, the DVL and IMU often must be placed in different locations and orientations on the AUV. Although ideally the DVL would be directly integrated with the IMU and precisely aligned during manufacture, this is rarely feasible in practical AUV implementations. \\
\noindent
\noindent
Various approaches have been developed to address this challenge. The early methods focused primarily on aligning the heading angle using least squares approaches \cite{joyce1989situ,pollard1989method}, which proved insufficient for high-precision navigation requirements. 
Kinsey et al. \cite{kinsey2007adaptive,kinsey2007situ} proposed using long baseline (LBL) acoustic positioning information with least squares techniques to estimate the alignment of the three degree of freedom alignment. Although effective, such methods require an external infrastructure with limited operational range. \\
\noindent
To address these limitations, Troni et al. \cite{troni2010new,troni2012field} developed self-contained alignment approaches that use only INS and DVL sensors onboard, demonstrating robust performance under deep-sea conditions despite real-world sensor noise. Their acceleration-based self-calibration methods eliminated the dependency on external sensors but required complex vehicle maneuvers and specific trajectories for effective calibration.\\
\noindent
A Kalman-filter-based online estimation approach \cite{zhaopeng2011online} addressed DVL alignment angle calibration during navigation. However, the method requires prolonged GNSS availability for convergence and exhibits degraded performance during straight-line constant-speed trajectories typical of AUV operations. \\
\noindent
A singular value decomposition (SVD) approach \cite{li2015alignment} reformulated the alignment problem as a least squares estimation using SINS/GNSS navigation data and DVL measurements as point sets. However, this method exhibits sensitivity to the selection of data window length selection and requires careful tuning of point set sizes to achieve optimal calibration performance.\\
\noindent
Li et al. \cite{li2022calibration} addressed the alignment problem using particle swarm optimization within a Wahba problem framework, achieving improved accuracy under dynamic conditions. However, their PSO-based approach introduces substantial computational complexity and convergence-time requirements that limit real-time applicability. \\
\noindent
Recent advances in deep learning have opened new avenues for addressing the challenges of sensor fusion and navigation \cite{chen2024deep, cohen2024inertial}. Deep learning models have demonstrated remarkable capabilities in pattern recognition and modeling of non-linear relationship from high-dimensional data. Successful applications in underwater navigation and engineering include DVL missing beam reconstruction \cite{cohen2022beamsnet,yona2024missbeamnet}, INS/DVL fusion \cite{cohen2024inertial,he2023deep,wang2019novel}, uncertainty estimation \cite{cohen2025adaptive,levy2025adaptive}, reinforcement learning for motion planning and control \cite{yu2025reinforcement,hadi2022deep}, DVL-denied navigation \cite{topini2023experimental}, and anomaly detection and recovery \cite{chen2025integrated}.
\noindent
In our previous work \cite{damari2025data}, we introduced AlignNet, a novel data-driven approach to INS/DVL alignment using a basic 1D convolutional neural network architecture. We demonstrated through simulation evaluation that AlignNet could achieve significant improvements in convergence time while maintaining accuracy comparable to traditional methods. However, the initial architecture faced limitations when transitioning from controlled simulation environments to real-world operational conditions, particularly with respect to feature extraction robustness and noise resilience under actual sensor measurement variations encountered in marine environments. \\
\noindent
This paper extends our previous research by presenting ResAlignNet, an enhanced deep learning framework that overcomes these limitations and other model-based shortcomings. We present comprehensive experimental validation using real-world navigation data collected during sea trials with the Snapir AUV, carried out in the Mediterranean Sea near Haifa, Israel, demonstrating practical applicability beyond simulation studies. We provide a detailed performance analysis of different sensor quality grades within a practical alignment range. This establishes the robustness of the approach under diverse operational conditions.
Our key contributions are:
\noindent
\begin{enumerate}
    \item \textbf{ResAlignNet}: A data-driven approach to the INS/DVL alignment problem using an advanced 1D ResNet-18 architecture specifically optimized for temporal sensor fusion, incorporating residual connections that enable deeper feature learning while maintaining gradient flow and improving convergence characteristics.
    \item \textbf{Self-Alignment}: Our proposed approach requires only the INS and DVL sensors without any other external sensors (such as GNSS). It does not require complex vehicle maneuvers or specific trajectories to achieve effective alignment.
    \item \textbf{Rapid Convergence}: We show that ResAlignNet greatly reduces the required time for alignment, offering a real-world practical advantage and reducing the time to mission start.
    \item \textbf{Sim2Real}: We demonstrate simulation-to-reality (Sim2Real) transfer learning capabilities, evaluating the potential for training on synthetic data while deploying on operational sensor measurements, thus reducing the burden of extensive data collection.
\end{enumerate}
Experimental results demonstrate that ResAlignNet achieves alignment accuracy in 0.8° using only 25 seconds of data collection, representing a 65\% reduction in convergence time compared to standard velocity-based methods. The approach maintains consistent sub-degree performance across different sensor quality grades and trajectory patterns, while traditional model-based methods exhibit severe degradation exceeding 35° RMSE with tactical-grade sensors.
\noindent
This paper is organized as follows. Section II formulates the INS/DVL alignment problem and reviews the mathematical framework. Section III describes our enhanced ResAlignNet architecture and training methodology. Section IV presents simulation validation, including simulation setup and results that compare our approach to the baseline method. Section V details field experiments, covering experimental setup and real-world performance evaluation. Section VI concludes with implications for underwater navigation systems and future research directions.

\section{Problem Formulation}\label{prob_form_sec}

\subsection{DVL Velocity Estimation}\label{dvl_basic_eq}
The DVL transmits and receives acoustic beams to and from the ocean floor. Using the Doppler shift effect, the DVL is able to estimate the AUV velocity in the DVL frame, denoted as $\boldsymbol{v}_{AUV}^{d}$. The beam arrays are generally mounted using the $"+"$ or $"\times"$ configurations \cite{liu2018innovative}. In the $"\times"$ configuration, also known as the "Janus" configuration, the beams are horizontally orthogonal with direction vectors defined by \cite{cohen2022beamsnet, braginsky2020correction}:
\begin{equation}\label{bi_in_h}
    \centering
        \boldsymbol{b}_{\dot{\imath}}=
        \begin{bmatrix} 
        \cos{\psi_{\dot{\imath}}}\sin{\alpha}\quad
        \sin{\psi_{\dot{\imath}}}\sin{\alpha}\quad
        \cos{\alpha}
    \end{bmatrix}_{1\times3}
\end{equation}
where, $\psi_{\dot{\imath}}$ and $\alpha$ are the yaw and pitch angles of the beam $i = 1,2,3,4$, respectively. For all beams, the pitch angle remains constant, whereas the yaw angle differs for each beam as \cite{yona2021compensating}:
\begin{equation}\label{yaw_of_beams}
    \centering
        \psi_{\dot{\imath}}=(\dot{\imath}-1)\cdot\frac{\pi}{2}+\frac{\pi}{4}\;[rad]\;,\; \dot{\imath}=1,2,3,4
\end{equation}
Stacking all beam directions (1) gives:
\begin{equation}\label{H_mat_defenition}
    \centering
        \mathbf{H}=
        \begin{bmatrix} \boldsymbol{b}_{1}\\\boldsymbol{b}_{2}\\\boldsymbol{b}_{3}\\\boldsymbol{b}_{4}\\
    \end{bmatrix}_{4\times3}
\end{equation}

Using $\mathbf{H}$, the AUV velocity vector in the DVL frame is connected to the measured velocity vector of the beams \cite{cohen2022beamsnet}:
\begin{equation}\label{simple_H_to_v_dvl}
    \centering
    \boldsymbol{v}_{beams} = \mathbf{H}\boldsymbol{v}^{d}_{DVL}
\end{equation}
where $\boldsymbol{v}_{beams} \in \mathbb{R}^4$ is the beam velocity vector and $\boldsymbol{v}^{d}_{AUV} \in \mathbb{R}^3$ is the AUV velocity vector in the DVL frame. To emulate real-world conditions, an error model is applied to each beam measurement \cite{yampolsky2025dcnet}, \cite{tal2017inertial}:
\begin{equation}\label{y_as_func_of_H_and_v_error_model}
    \centering
        \boldsymbol{\tilde{v}}_{beams} = [\mathbf{H} \boldsymbol{v}_{DVL}^{d}(1+\boldsymbol{s}_{DVL})]+\boldsymbol{b}_{DVL}+\boldsymbol{\sigma}_{DVL}
\end{equation}
where $\boldsymbol{s}_{DVL}$ is the DVL scale factor, $\boldsymbol{b}_{DVL}$ is the DVL bias, and $\boldsymbol{\sigma}_{DVL}$ is the DVL Gaussian distributed zero mean white noise.

The AUV velocity vector is obtained by minimizing the following cost.
\begin{equation}\label{ls_form_eq}
    \centering
    \tilde{\boldsymbol{v}}_{DVL}^{d} = \underset{\boldsymbol{{v}}_{DVL}^{d}}{\argmin}{||\boldsymbol{\tilde{v}}_{beams}-\mathbf{H}\boldsymbol{{v}}_{DVL}^{d} ||}^{2}
\end{equation}
yielding a least-squares solution:
\begin{equation}\label{psudo_inverse}
    \centering
    \tilde{\boldsymbol{v}}_{DVL}^{d} = (\mathbf{H}^{T}\mathbf{H})^{-1}\mathbf{H}^{T}\boldsymbol{\tilde{v}}_{beams}
\end{equation}

\subsection{INS Equations of Motion}\label{ins_equations}
The INS process the accelerometer and gyroscope to estimate the velocity vector required for the INS/DVL alignment process.
Working with low-cost inertial sensors, certain simplifications can be made to the navigation equations of motion. Specifically, the Earth's rotation rate and transport rate effects can be neglected. Under these assumptions, the fundamental INS equations of motion are given by \cite{titterton2004strapdown}:
\begin{equation}
    \centering
    \dot{\boldsymbol{p}}^{n} = \boldsymbol{v}^{n}
    \label{eq:p_n}
\end{equation}
\begin{equation}
    \centering
    \dot{\boldsymbol{v}}^{n} = \mathbf{R}^{n}_{b}\boldsymbol{f}^{n}_{ib} + \boldsymbol{g}^{n}
    \label{eq:v_n}
\end{equation}
\begin{equation}
    \centering
    \dot{\mathbf{R}}^{n}_{b} = \mathbf{R}^{n}_{b}\boldsymbol{\Omega}^{b}_{ib}
    \label{eq:R}
\end{equation}
where ${\boldsymbol{f}}^{n}_{ib}$ is the specific force vector expressed in the local navigation frame, (North-East-Down, NED), $\boldsymbol{p}^{n}$ is the position vector expressed in the local navigation frame, $\boldsymbol{v}^{n}$ is the velocity vector expressed in the local navigation frame, $\mathbf{R}^{n}_{b}$ is the transformation matrix from body to local navigation frame, $\boldsymbol{g}^{n}$ is the gravity vector expressed in the local navigation frame, and $\boldsymbol{\Omega}^{b}_{ib}$ is the skew-symmetric matrix formed from gyroscope measurements, defined as:
\begin{equation}
    \centering
    \boldsymbol{\Omega}^{b}_{ib} = 
    \begin{bmatrix}
    0 & -\omega_z & \omega_y \\
    \omega_z & 0 & -\omega_x \\
    -\omega_y & \omega_x & 0
    \end{bmatrix}
\end{equation}

A typical error model for the specific force vector is \cite{groves2013principles}:
\begin{equation}
    \centering
    \tilde{\boldsymbol{f}}^{b}_{ib} = \boldsymbol{S}_a\boldsymbol{f}^{b}_{ib} + \boldsymbol{b}_a + \boldsymbol{\sigma}_a
\end{equation}
where $\tilde{\boldsymbol{f}}^{b}_{ib}$ is the measured specific force vector, $\boldsymbol{S}_a$ is a diagonal scale factor matrix, $\boldsymbol{b}_a$ is the bias vector, and $\boldsymbol{\sigma}_a$ is Gaussian white noise of zero mean. \\
\noindent
Similarly, for the angular rate measurements:
\begin{equation}
    \centering
    \tilde{\boldsymbol{\omega}}^{b}_{ib} = \boldsymbol{S}_g\boldsymbol{\omega}^{b}_{ib} + \boldsymbol{b}_g + \boldsymbol{\sigma}_g
\end{equation}
where $\tilde{\boldsymbol{\omega}}^{b}_{ib}$ is the measured angular velocity, $\boldsymbol{S}_g$ is the scale factor matrix, $\boldsymbol{b}_g$ is the bias vector and $\boldsymbol{\sigma}_g$ is the Gaussian white noise with zero mean.

\subsection{The INS/DVL Alignment Problem}

The INS/DVL alignment problem can be formulated as the estimation of the rotation matrix $\mathbf{C}_d^b \in SO(3)$ between the body frame $(b)$ and the DVL frame $(d)$, assuming that the sensitive axes of the inertial sensors align with the body frame. \\
\noindent
A standard In Situ approach to the alignment of the INS/DVL sensors is based on a velocity-based method that integrates the estimated inertial acceleration data and compares it with velocity measurements from the DVL \cite{troni2012field}. \\
\noindent
The velocity vector expressed in the body frame is obtained by:
\begin{equation}\label{velocity_based_eq}
    \centering
    \boldsymbol{v}^{b}(t) = \mathbf{R}_{i}^{b}(t)^{T}\int_{t_0}^{t}\mathbf{R}_{i}^{b}(\tau)\boldsymbol{a}_{ib}^{b}(\tau)d\tau + \boldsymbol{v}^{b}(t_0)
\end{equation}
\noindent
where $\mathbf{R}_{i}^{b}$ is the rotation matrix from the inertial frame to the body frame, $\boldsymbol{a}_{ib}^{b}$ estimated inertial acceleration using \ref{eq:p_n}-\ref{eq:R}, and $\boldsymbol{v}^{b}(t_0)$ is the initial velocity in the body frame, which is assumed to be known. \\
\noindent
The estimated velocity, calculated from the inertial sensors measurements, can then be compared with the velocity measurements from the DVL:
\begin{equation}\label{velocity_based_eq}
    \centering
    \boldsymbol{v}^{b}(t) = \mathbf{R}_{b}^{d}\boldsymbol{v}^{d}(t)
\end{equation}
\noindent
The INS/DVL alignment problem is transformed into an optimization problem, which is known as the Wahba problem \cite{wahba1965least}. The cost function can be defined as the mean squared error of the residuals between these two velocity measurements on $N$ time instances $t_i$, $i = 1, 2, \ldots, N$.
\begin{equation}
\boldsymbol{C}(\mathbf{R}_{b}^{d}) = \frac{1}{N}\sum_{i=1}^{N} \left\|\boldsymbol{v}^{b}[t_i] - \mathbf{R}_{b}^{d}\boldsymbol{v}^{d}[t_i]\right\|^2
\end{equation}
\noindent
The optimal linear least-squares estimate for $\mathbf{R}_{b}^{d}$ is determined by solving the following optimization problem:
\begin{equation}
\hat{\mathbf{R}}_{b}^{d} = \argmin_{\mathbf{R}_{b}^{d} \in SO(3)} \boldsymbol{C}(\mathbf{R}_{b}^{d})
\end{equation}
\noindent
Several computational approaches exist to solve the Wahba problem \cite{eggert1997estimating}. The SVD-based solution stands out among these techniques because of its numerical stability, good accuracy performance, and inherent preservation of SO(3) constraints for the transformation matrix. Given its proven reliability, we adopt the SVD-based approach as our baseline method to evaluate the proposed deep neural network method.
For a detailed description of the SVD-based solution, the reader is referred to \cite{Umeyama1991Least}.






\section{Proposed Approach}\label{enhanced_approach}
\begin{figure*}[t]
    \centering
    \includegraphics[width=1\textwidth]{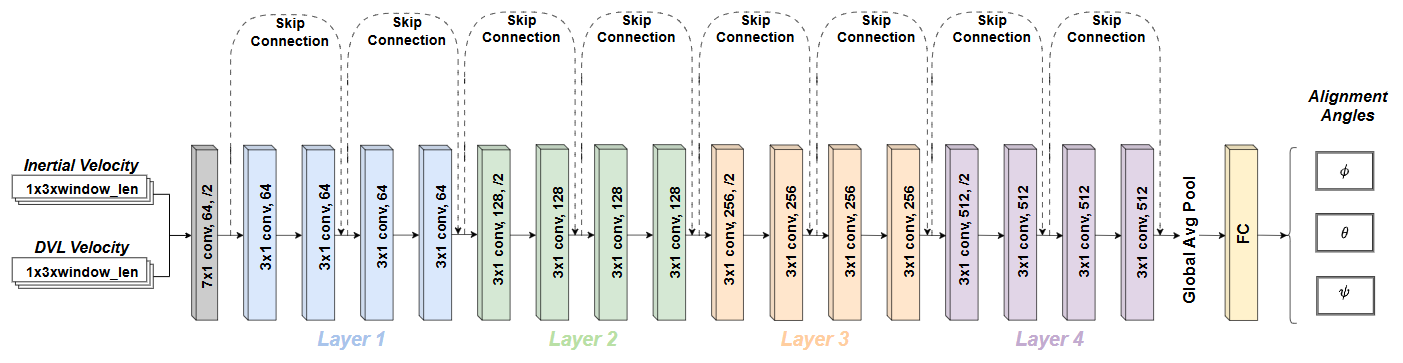}
    \caption{ResAlignNet architecture utilizing 1D ResNet-18 structure with residual connections for alignment parameters estimation.}
    \label{fig:network_architecture}
\end{figure*}

In this work, we propose the ResAlignNet framework for accurate INS/DVL alignment. ResAlignNet addresses the sensor alignment problem by transforming it into a supervised learning task that directly estimates rotation parameters between the inertial and Doppler velocity sensor reference frames. Our data-driven approach eliminates the convergence time limitations and motion pattern dependencies of standard model-based methods by learning complex nonlinear relationships from synchronized velocity measurements. The methodology operates in-situ using only onboard sensors, enhancing operational autonomy for underwater vehicles.\\
\noindent
ResAlignNet employs a modified 1D ResNet-18 architecture specifically optimized for temporal sensor data processing. This architectural design provides significant improvements in feature extraction capability, convergence characteristics, and robustness to sensor noise for INS/DVL alignment tasks. Figure \ref{fig:network_architecture} illustrates the complete architecture of our improved ResAlignNet framework. \\
\noindent
\noindent
\subsection{Input Layer and Feature Extraction}
The input layer accepts temporally aligned velocity measurements from both DVL and inertial sensors, combining them into a six-dimensional tensor that captures complete motion characteristics. This tensor has dimensions [$batch\_size$, $window\_size$, $6$], where $window\_size$ represents the temporal dimension of the measurement window, and the six dimensions correspond to three velocity components of each sensor. \\
\noindent
The sampling rate disparity between sensors is handled during data preparation, where DVL samples are organized into overlapping temporal windows with a one sample stride. For each DVL window, the temporally nearest IMU velocity samples are selected to form the corresponding measurement pairs. \\
\noindent
\subsection{Convolutional Layers}
The network employs 1D convolutional layers for extracting temporal features from the velocity measurement sequences. The architecture consists of an initial convolutional layer followed by four residual layers, each containing multiple convolutional operations. \\
\noindent
The initial convolutional layer applies 64 filters with a kernel size of 7 and a stride of 2, performing feature extraction while reducing the temporal dimension.\\
\noindent
The 1D convolution operation is mathematically defined as:
\begin{equation}
y[n] = \sum_{k=0}^{K-1} x[n + k] \cdot w[k] + b
\end{equation}
\noindent
where $x[n]$ represents the input sequence, $w[k]$ denotes the learnable filter weights, $K$ is the kernel size, $b$ is the bias term, and $y[n]$ is the output feature map.
\noindent
For multichannel processing with $C_{in}$ input channels and $C_{out}$ output channels:
\begin{equation}
y_j[n] = \sum_{i=0}^{C_{in}-1} \sum_{k=0}^{K-1} x_i[n \cdot s + k] \cdot w_{i,j}[k] + b_j
\end{equation}
\noindent
where $s$ represents the stride, $i$ indexes the input channels, $j$ indexes the output channels, and $w_{i,j}[k]$ are the channel-specific filter weights. \\
\noindent
\subsection{Residual Blocks}
The network incorporates four residual layers (Layers 1-4, in Figure \ref{fig:network_architecture}), each containing multiple residual blocks that enable deep feature learning while mitigating the problem of vanishing gradient. Each residual block implements skip connections that allow information to bypass convolutional operations, facilitating training of deeper networks.\\
\noindent
The fundamental residual block operation is defined as:
\begin{equation}
\mathbf{y} = \mathcal{F}(\mathbf{x}, \{W_i\}) + \mathbf{x}
\end{equation}
\noindent
where $\mathbf{x}$ is the input, $\mathcal{F}(\mathbf{x}, \{W_i\})$ represents the residual mapping learned by the stacked convolutional layers with weights $\{W_i\}$, and $\mathbf{y}$ is the output. The skip connection adds the input directly to the processed features, enabling gradient flow through the network. \\
\noindent
The four residual layers progressively increase in complexity with channel dimensions of 64, 128, 256, and 512 respectively. Each layer contains multiple residual blocks with 3×1 convolutional kernels, batch normalization, and ReLU activation functions. Skip connections preserve gradient flow during backpropagation, enabling effective training of the deep architecture while extracting hierarchical temporal features essential for accurate alignment parameter estimation. \\

\subsection{Feature Aggregation and Output Layers}
The final phase of the network performs feature aggregation and alignment parameter prediction through global pooling and fully connected layers. Following residual blocks, a global average pooling layer reduces the temporal dimension by computing the mean across all time steps, transforming the feature maps into a fixed-size representation. \\
\noindent
The pooled features are then fed into a fully connected (FC) layer that maps the high-dimensional feature representation to the three alignment angles (roll $\phi$, pitch $\theta$, and yaw $\psi$), representing the rotation parameters between the IMU and DVL coordinate frames.

\subsection{Loss Function and Optimization}
The network is trained by minimizing the mean squared error (MSE) between the predicted alignment angles and the GT values. The loss function is computed across the three Euler angles that define the rotation between the IMU and the DVL reference frames:
\noindent
\begin{equation}\label{mse_eq}
\text{MSE}(\boldsymbol{\alpha}, \hat{\boldsymbol{\alpha}}) =  \frac{\sum_{i=1}^{N} \sum_{j \in \{\phi,\theta,\psi\}}(\boldsymbol{\alpha}_{i,j} - \hat{\boldsymbol{\alpha}}_{i,j})^{2}}{N}
\end{equation}
\noindent
where $\boldsymbol{\alpha}$ represents the Euler GT alignment angles, $\hat{\boldsymbol{\alpha}}$ represents the network-estimated angles, $N$ is the number of training samples, and $j$ indexes the three rotational components (roll $\phi$, pitch $\theta$, and yaw $\psi$). \\
\noindent
The network is optimized using Adam optimizer with an initial learning rate of $10^{-7}$ and a batch size of 32.

\section{Simulation Validation}\label{res_sec}
Our evaluation employs a simulation-based analysis that serves two key purposes. First, the simulation environment allows us to systematically examine the limitations and performance characteristics of current model-based alignment approaches under controlled conditions with known GT parameters. Second, the simulated data generated in this phase provides a foundation for a subsequent Sim2Real transfer learning analysis, where models trained exclusively on synthetic data are evaluated against real-world sensor measurements. This dual-purpose simulation framework establishes both a performance baseline for comparison and a practical pathway toward reducing the dependency on extensive real-world data collection during model development.
\noindent
\subsection{Simulation Setup}\label{simulation_setup}
\begin{figure}[!h]
    \centering    \includegraphics[width=\columnwidth]{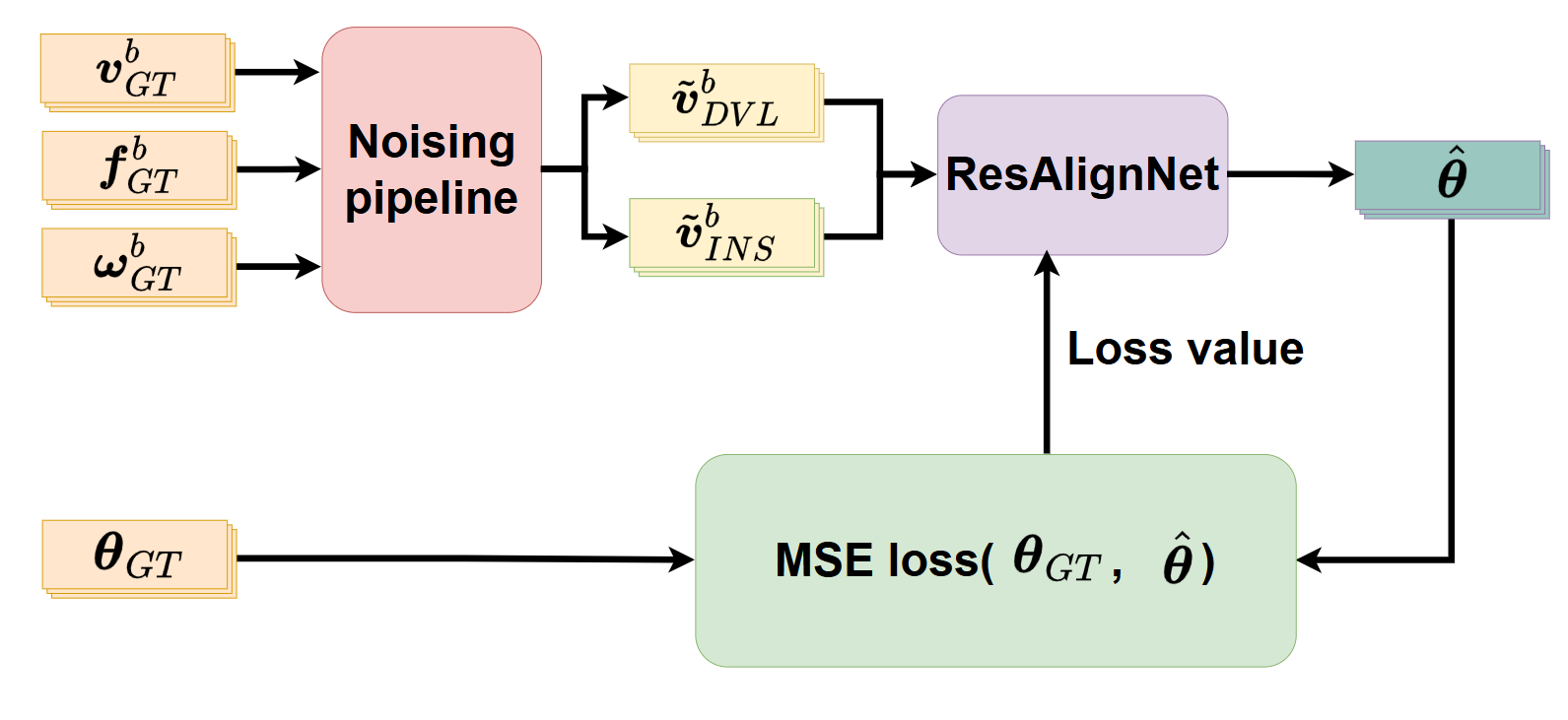}
    \caption{Overview of the ResAlignNet training pipeline. The GT trajectory data including velocity vector ($\boldsymbol{v}^b_{GT}$), specific force vector ($\boldsymbol{f}^b_{GT}$), and angular velocity rate vector ($\boldsymbol{\omega}^b_{GT}$) is processed through a noising pipeline to generate simulated DVL and INS velocities, which serve as input to ResAlignNet. The network estimates the alignment angles ($\hat{\boldsymbol{\theta}}$), which are compared against the GT alignment angles ($\boldsymbol{\theta}_{GT}$) using MSE loss to train the network.}
    \label{fig:sim_pipeline}
\end{figure}

\begin{figure}[!h]
    \centering    \includegraphics[width=\columnwidth]{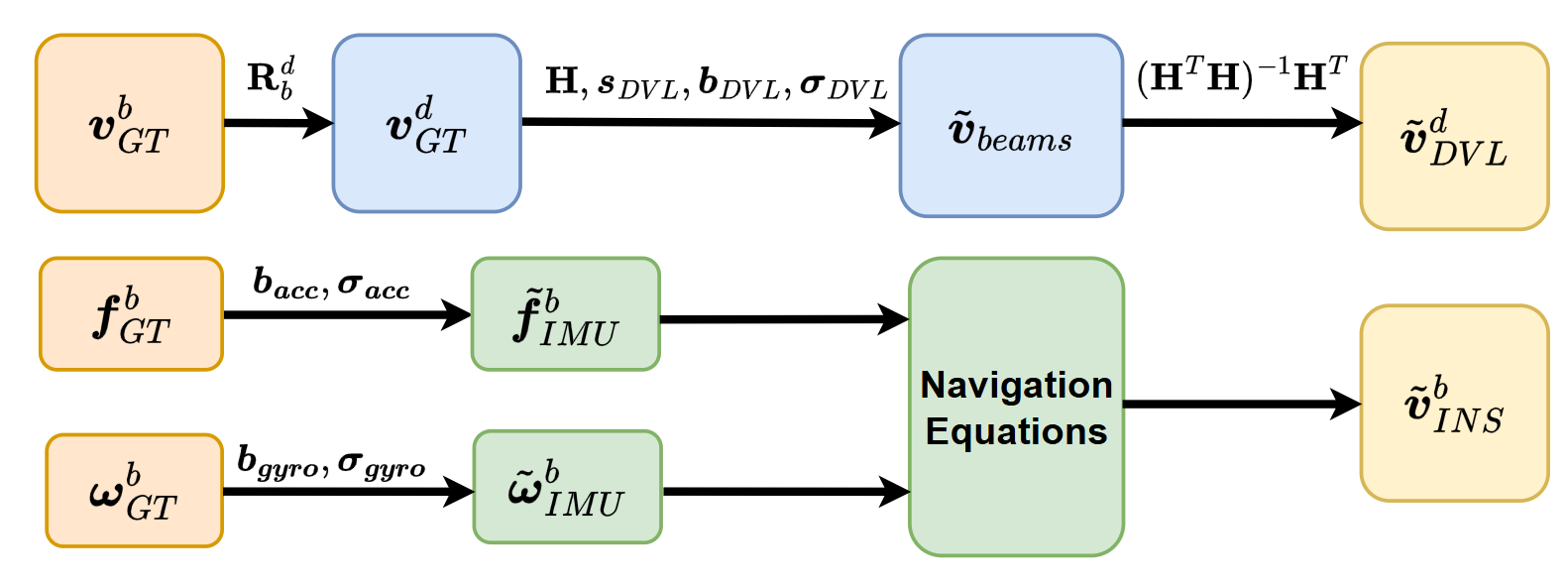}
    \caption{Detailed noising pipeline for simulation data generation. GT velocity vector in the body frame ($v^b_{GT}$) is transformed to the DVL frame using the generated GT alignment parameters ($\mathbf{C}_d^b$) and processed through the DVL error model incorporating scale factors, biases, and zero mean white Gaussian noise. Simultaneously, the inertial readings GT are processed through the INS error model with accelerometer and gyroscope errors (biases $\boldsymbol{b}_a$, $\boldsymbol{b}_g$, and zero mean white Gaussian noise $\boldsymbol{\sigma}_a$, $\boldsymbol{\sigma}_g$) before integration through the equations of motion to produce noisy INS velocity vector estimates ($\tilde{\boldsymbol{v}}^b_{INS}$).}
    \label{fig:noising_pipeline_overview}
\end{figure}

We developed a comprehensive MATLAB-Simulink simulation framework to evaluate our proposed approach against the model-based baseline. Our simulation environment incorporates a six-degree-of-freedom (6-DOF) AUV dynamics model with realistic hydrodynamic forces, detailed sensor error models, and comprehensive sensor error characteristics. The simulation framework begins with the generation of GT trajectory data, including position, velocity, and orientation parameters over time.\\
\noindent
\noindent
Figure \ref{fig:sim_pipeline} presents the overall simulation training pipeline of our proposed approach. The pipeline begins with the simulative GT trajectory data including velocity ($\boldsymbol{v}^b_{GT}$), specific force ($\boldsymbol{f}^b_{GT}$), and angular velocity rate ($\boldsymbol{\omega}^b_{GT}$) from the simulated trajectory that is processed through a noising pipeline (detailed in Figure \ref{fig:noising_pipeline_overview}) to simulate realistic sensor measurements. For sensor error modeling, we implement comprehensive error characteristics for both DVL and INS measurements. The DVL error model incorporates scale factors, biases, and zero mean white Guassain noise. Similarly, the INS error model includes biases and noise for both accelerometer and gyroscope measurements, as described in Section \ref{ins_equations}. These noisy measurements from both DVL and INS are then fed into ResAlignNet, which estimates the alignment angles between the sensor frames. The network is trained by comparing these estimated angles with GT alignment angles using the MSE loss, with the loss value feeding back to optimize the network parameters.
\noindent
\noindent
\begin{figure}[!h]
    \centering    \includegraphics[width=\columnwidth]{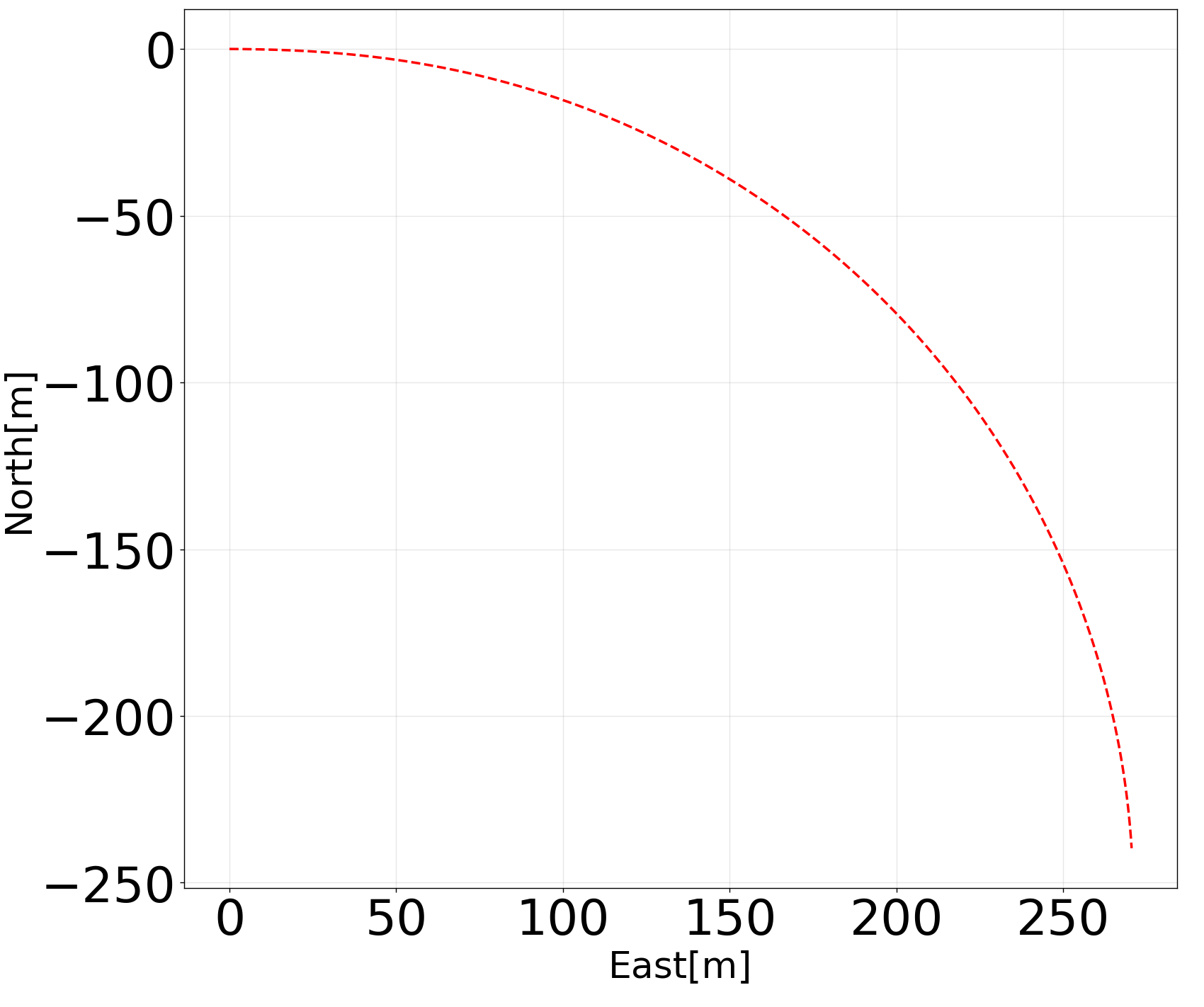}
    \caption{Simulated 200-second right-turn trajectory at constant speed of 2m/s.}
    \label{fig:sim_traj}
\end{figure}
\noindent
The trajectory examined is shown in Figure \ref{fig:sim_traj}. This fundamental maneuver represents a common
operational scenario in underwater navigation that provides
sufficient dynamic excitation for the estimation of the alignment
parameters. The trajectory spans a 200-second interval to ensure adequate data collection for convergence analysis.\\
\noindent
We systematically vary the INS/DVL alignment to assess the robustness of the algorithm in realistic deployment scenarios. The alignment range examined consists of Euler angles between 0-5 degrees per axis, representing typical installation tolerances encountered in operational AUV systems.
\noindent
In addition, we evaluate performance across two IMU grades to assess sensor-grade dependency. Navigation-grade IMU parameters feature superior accuracy, stability, and reduced drift characteristics, while tactical-grade IMU parameters represent moderate-performance specifications that match those of our experimental AUV platform. This dual-grade evaluation ensures that our approach generalizes across different sensor quality levels commonly deployed in underwater vehicles. The complete simulation parameters for both configurations are detailed in Table \ref{tab:simulation_parameters}.\\
\noindent
The generated dataset comprises 9,826 individual recordings, each with a duration of 200 seconds, resulting in a total dataset duration of 545 hours. For each recording, we systematically vary the INS/DVL alignment across the three Euler angles (roll, pitch, and yaw) within the 0-5 degree range, generating 4,913 distinct alignment configurations per IMU specification type. The dataset includes both navigation-grade and tactical-grade IMU specifications (4,913 configurations each), enabling comprehensive evaluation across different sensor quality levels. The complete dataset follows standard machine learning protocols with 60\% allocated for training, 20\% for validation, and 20\% for testing. This partitioning ensures robust statistical evaluation while providing sufficient training samples for the neural network approach across diverse alignment scenarios and sensor quality levels.
\noindent
\begin{table}[!h]
\caption{Simulation, DVL, and IMU parameters used in this study.}
\label{tab:simulation_parameters}
\centering
\adjustbox{width=\columnwidth,center}
{
\begin{tabular}{|l|c|}
\hline
\multicolumn{2}{|c|}{\textbf{Simulation Parameters}} \\
\hline
\textbf{Parameter} & \textbf{Value} \\
\hline
Time duration & 200 (s) \\
AUV velocity & 2 (m/s) \\
\hline
\multicolumn{2}{|c|}{\textbf{DVL Parameters}} \\
\hline
\textbf{Parameter} & \textbf{Value} \\
\hline
DVL rate & 5 (Hz) \\
DVL noise & 0.008 (m/s) \\
DVL bias & 0.001 (m/s) \\
DVL scale factor & 0.5 (\%) \\
\hline
\multicolumn{2}{|c|}{\textbf{IMU Parameters}} \\
\hline
\textbf{Parameter} & \textbf{Navigation Grade IMU / Tactical Grade IMU} \\
\hline
IMU rate & \multicolumn{1}{c|}{100 (Hz)} \\
Accelerometer bias & 0.1 (mg) / 1 (mg) \\
Gyro bias & 1 (°/h) / 10 (°/h) \\
Accelerometers noise & 0.001 (mg/$\sqrt{Hz}$) / 0.01 (mg/$\sqrt{Hz}$) \\
Gyro noise & 0.01 (°/$\sqrt{h}$) / 0.1 (°/$\sqrt{h}$) \\
\hline
\end{tabular}
}
\end{table}

\subsection{Evaluation Metrics}\label{evaluation_metrics}
To comprehensively assess alignment estimation performance, we employ two complementary metrics that quantify accuracy from different perspectives.

\subsubsection{Root Mean Square Error (RMSE)}
The RMSE quantifies the precision of the alignment estimation in the Euler angle space. It is defined as:
\begin{equation}\label{rmse_eq}
\text{RMSE}(\boldsymbol{\alpha}, \hat{\boldsymbol{\alpha}}) = \sqrt{ \frac{\sum_{i=1}^{N} \sum_{j \in \{\phi,\theta,\psi\}}(\boldsymbol{\alpha}_{i,j} - \hat{\boldsymbol{\alpha}}_{i,j})^{2}}{N}}
\end{equation}
where $N$ is the number of test samples, $\boldsymbol{\alpha}_{i,j}$ denotes the GT alignment angle for sample $i$ and angle component $j$, and $\hat{\boldsymbol{\alpha}}_{i,j}$ represents the predicted alignment angle, with $j \in \{\phi,\theta,\psi\}$ corresponding to the roll, pitch, and yaw angles respectively.\\

\subsubsection{Average Orientation Error (AOE)}
The AOE provides a rotation matrix-based assessment of alignment accuracy, directly measuring the angular deviation in SO(3) space. It is computed as:
\begin{equation}\label{aoe_eq}
\text{AOE} = \sqrt{\sum_{n=1}^{M} \frac{1}{M}\left\|\log\left(\mathbf{R}_n^T\hat{\mathbf{R}}_n\right)\right\|_2^2}
\end{equation}
where $M$ is the number of test samples, $\mathbf{R}_n$ is the GT rotation matrix from the body frame to the DVL frame, $\hat{\mathbf{R}}_n$ is the estimated rotation matrix, and $\log(\cdot)$ denotes the logarithm map from SO(3) to its Lie algebra $\mathfrak{so}(3)$.

\subsection{Simulation Results}\label{simulation_results}
\begin{figure}[!h]
    \centering
    \includegraphics[width=\columnwidth]{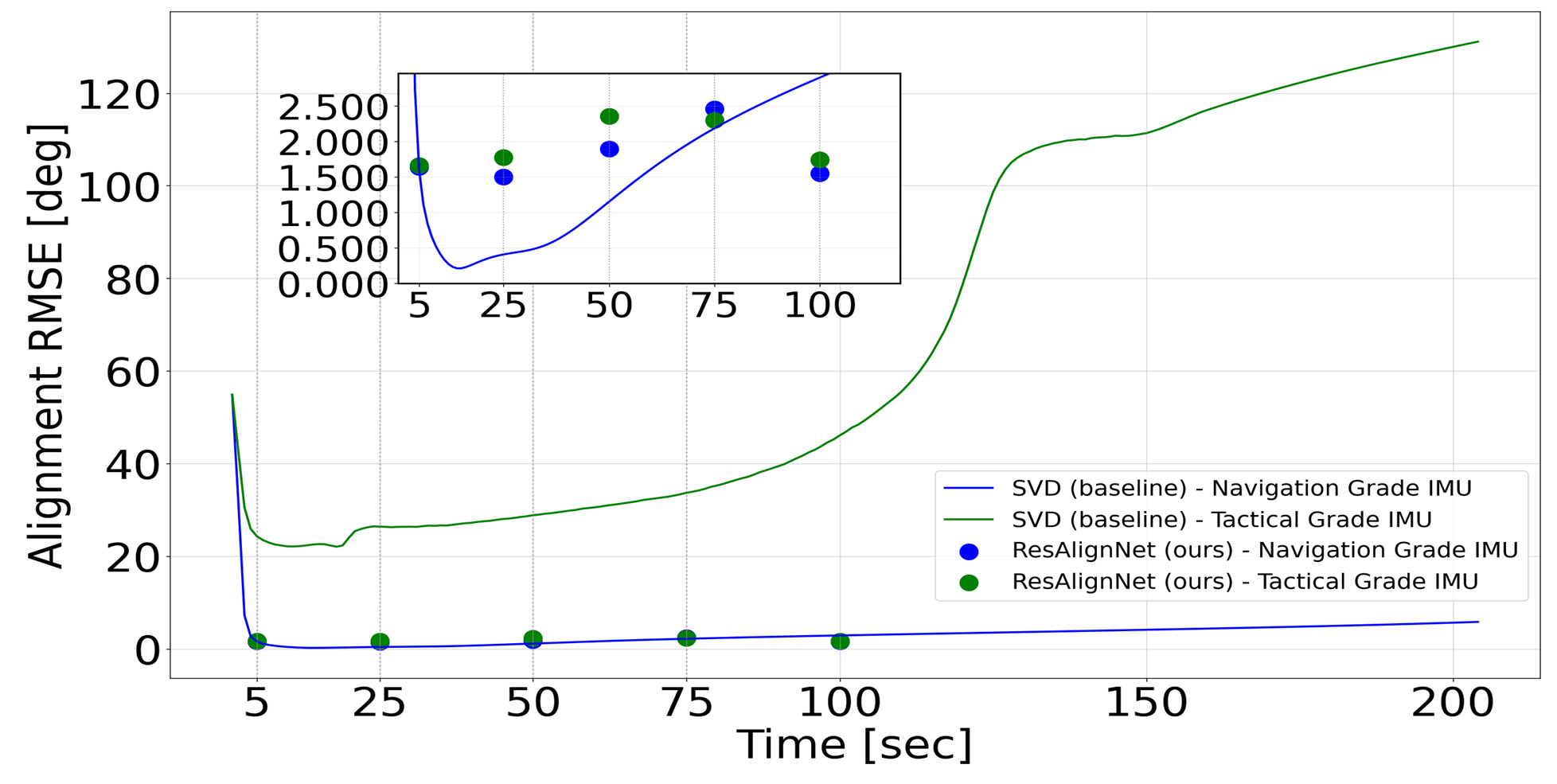}
    \caption{Simulation results: RMSE alignment performance for an alignment range of 0-5 degrees comparing ResAlignNet and SVD methods for turn trajectory using tactical-grade IMU.}
    \label{fig:small_misalignment_results}
\end{figure}

\begin{figure}[!h]
    \centering
    \includegraphics[width=\columnwidth]{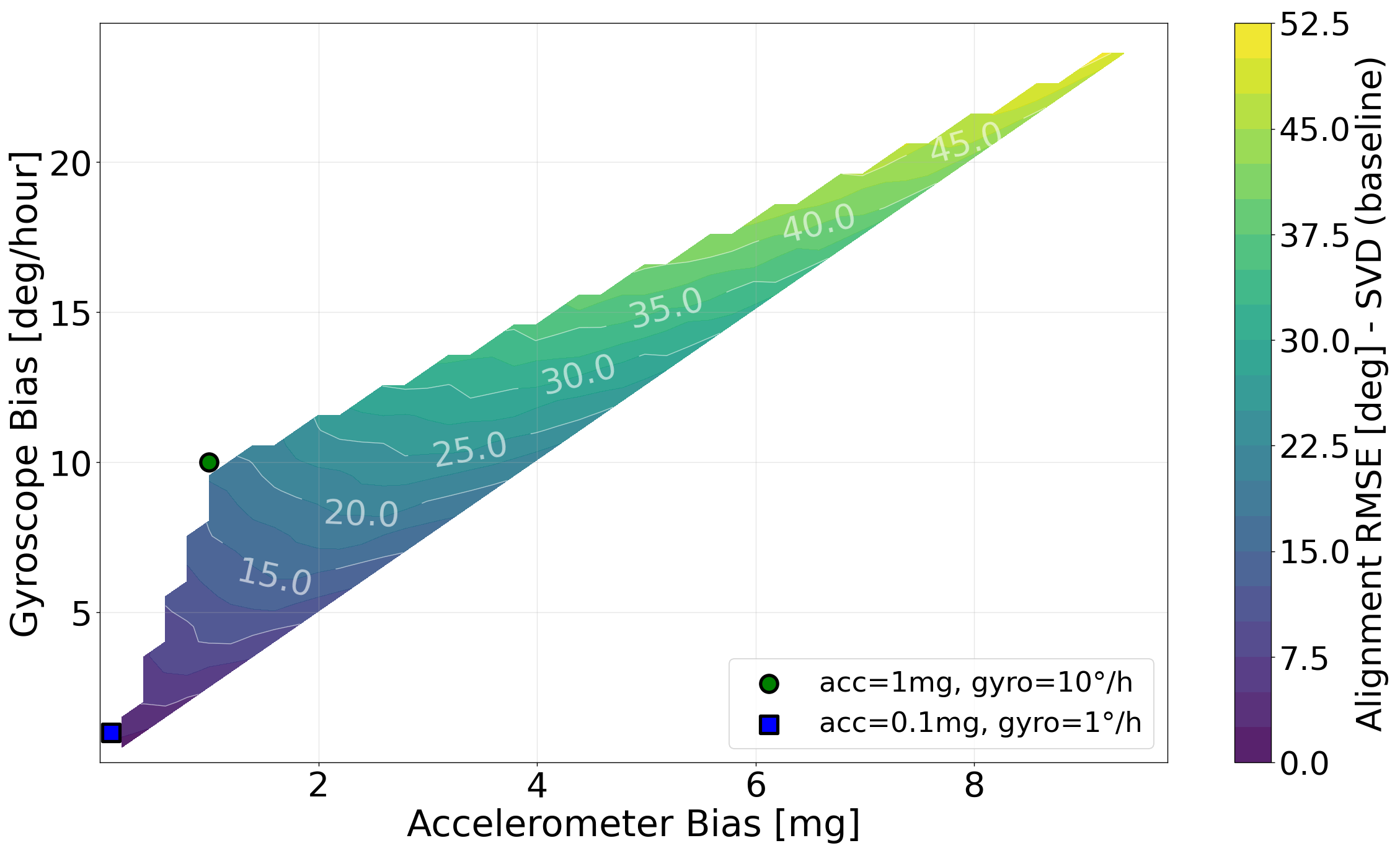}
    \caption{SVD method alignment RMSE sensitivity to IMU sensor biases for the simulated turn trajectory. The contour plot shows the minimum achievable alignment RMSE across different accelerometer and gyroscope bias combinations over a 200-second evaluation window. The two markers indicate navigation-grade (blue square) and tactical-grade (green circle) IMU specifications used in this study. Results demonstrate substantial performance degradation with increasing sensor bias magnitudes.}
    \label{fig:RMSE_bias}
\end{figure}

We evaluated the performance of ResAlignNet against the SVD-based baseline for the turn trajectory (shown in Figure \ref{fig:sim_traj}) in different IMU grades. The results are presented for the alignment range of 0-5 degrees.\\
\noindent
Figure \ref{fig:small_misalignment_results} presents the alignment performance of the RMSE for the turn trajectory under the alignment conditions examined. The results demonstrate different performance characteristics between the two approaches in different sensor qualities.\\
\noindent
ResAlignNet maintains consistent performance in both IMU grades, achieving RMSE values between 1.5-2.5 degrees in all evaluation windows. The turning trajectory demonstrates improved SVD performance with navigation-grade sensors, achieving optimal results of 0.42 degrees at 15 seconds, although performance varies considerably between different window sizes. With tactical-grade IMU, SVD performance remains poor for turning maneuvers, with RMSE values exceeding 23 degrees throughout all evaluation periods. ResAlignNet shows minimal sensitivity to sensor quality, consistently delivering sub-2.5 degree accuracy regardless of operational conditions.\\
\noindent
To investigate the sensitivity of the SVD-based alignment method to sensor error characteristics, we performed a systematic analysis examining performance across varying accelerometer and gyroscope bias magnitudes. Figure \ref{fig:RMSE_bias} presents a contour graph that illustrates the minimum achievable alignment RMSE for the SVD method as a function of accelerometer bias (ranging from 0.1 to 10 mg) and gyroscope bias (ranging from 1 to 25 deg/hour) using the same turn trajectory evaluated in Section \ref{simulation_results}.\\
\noindent
The analysis reveals a strong correlation between the magnitude of sensor bias and alignment accuracy. The contour plot demonstrates that SVD performance decreases significantly as the bias values increase, with the minimum achievable RMSE growing approximately linearly with both accelerometer and gyroscope bias. For navigation-grade IMU specifications (accelerometer bias = 0.1 mg, gyroscope bias = 1°/h, marked with a blue square in Figure \ref{fig:RMSE_bias}), the SVD method achieves a minimum RMSE of approximately 0.42 degrees. However, at tactical-grade specifications (accelerometer bias = 1 mg, gyroscope bias = 10°/h, marked with a green circle in Figure \ref{fig:RMSE_bias}), the minimum RMSE increases to approximately 23 degrees.\\
\noindent
\noindent

\section{Field Experiments}\label{res_sec}

\subsection{Experimental Setup}
The experimental validation of our proposed approach utilizes real-world navigation data collected during sea trials conducted in the Mediterranean Sea near Haifa, Israel. The trials were carried out using Snapir, a modified ECA Group A18D midsize autonomous underwater vehicle specifically configured for deep-water research operations \cite{eca_group}. Snapir is capable of autonomous mission execution at depths up to 3000 meters with an operational endurance of 21 hours, making it well-suited for extended underwater navigation studies.\\
\noindent
Snapir incorporates a comprehensive navigation sensor suite designed for precision underwater operations. The vehicle is equipped with an iXblue Phins Subsea \cite{ixblue_phins}, a fiber optic gyroscope based high-performance subsea inertial navigation system. Velocity measurements are provided by a Teledyne RDI Work Horse Navigator Doppler velocity log \cite{teledyne_dvl}, which achieves accurate velocity measurements with a standard deviation of 0.02 m/s. The temporal characteristics of these sensors align with typical AUV configurations, with the INS operating at 100 Hz and the DVL at 1 Hz sampling rates. The dataset used in this study was collected during sea trials and contains various operational scenarios, including different maneuvers, depths, and speeds as addressed below \cite{cohen2025adaptive}. 

\subsection{Experimental Dataset}
\begin{figure}
    \centering
    \includegraphics[width=\columnwidth]{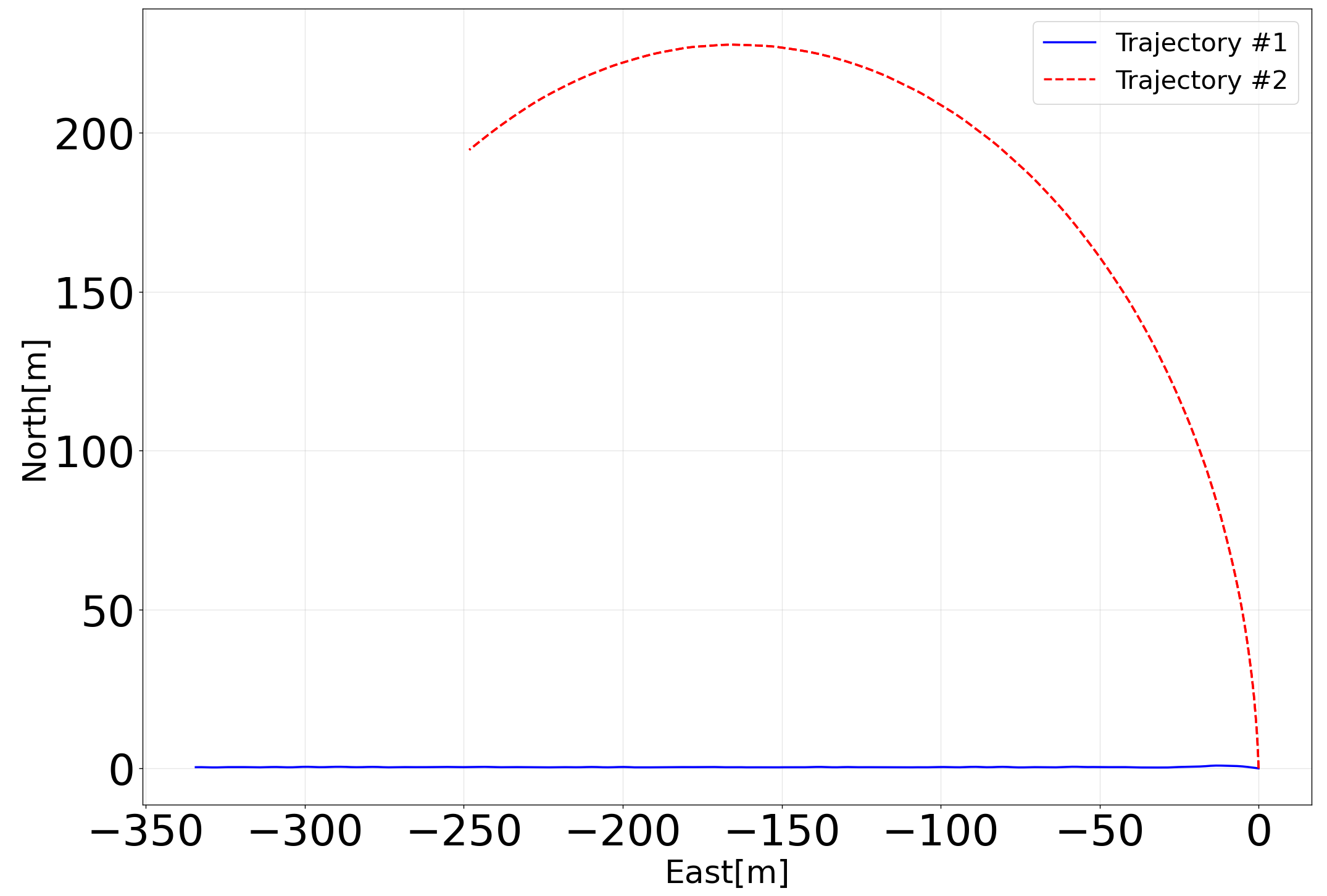}
    \caption{AUV Snapir real trajectories: Trajectory \#1 (straight line) and Trajectory \#2 (long turn) as viewed from above.}
    \label{fig:real_data_trajectories}
\end{figure}

To generate sufficient training diversity from limited real-world deployments where the AUV maintains fixed sensor alignment, we employ a systematic data augmentation strategy. Multiple alignment scenarios are synthetically created by applying diverse rotation matrices to the original DVL sensor measurements, effectively generating a comprehensive range of alignment conditions while preserving authentic sensor characteristics and noise profiles.\\
\noindent
The alignment range examined consists of Euler angles between 0-5 degrees per axis, representing typical installation tolerances in operational AUV systems. For each trajectory segment, we generate multiple samples by systematically varying the alignment parameters across this range. The original collected data comprises two trajectory segments, each lasting 200 seconds. Through the data augmentation process, where multiple alignment scenarios are synthetically created for each trajectory segment, the total augmented dataset duration reaches 545 hours. The augmented dataset follows standard machine learning protocols with 60\% allocated for training, 20\% for validation, and 20\% for testing.\\
\noindent
For performance evaluation, two representative trajectory patterns were selected from the sea trial data, as shown in Figure \ref{fig:real_data_trajectories}. Trajectory \#1 consists of a straight-line segment representing steady-state navigation, while Trajectory \#2 involves a long-turn maneuver providing dynamic motion conditions. The specifications of both trajectories are detailed in Table \ref{tab:trajectory_specifications}.

\begin{table}[!h]
\caption{Experimental test trajectories specifications.}
\label{tab:trajectory_specifications}
\centering
\begin{tabular}{|c|c|c|c|c|}
\hline
\textbf{Trajectory} & \textbf{Pattern} & \textbf{Duration} & \textbf{Avg Velocity} & \textbf{Avg Depth} \\
\textbf{\#} & & \textbf{(s)} & \textbf{(m/s)} & \textbf{(m)} \\
\hline
1 & Straight line & 200 & 2.05 & -10.61 \\
\hline
2 & Long turn & 200 & 1.83 & -15 \\
\hline
\end{tabular}
\end{table}
\noindent

\subsection{Experimental Results}\label{experimental_results}
\begin{figure}[!h]
    \centering
    \begin{subfigure}{\columnwidth}
        \centering
        \includegraphics[width=\textwidth]{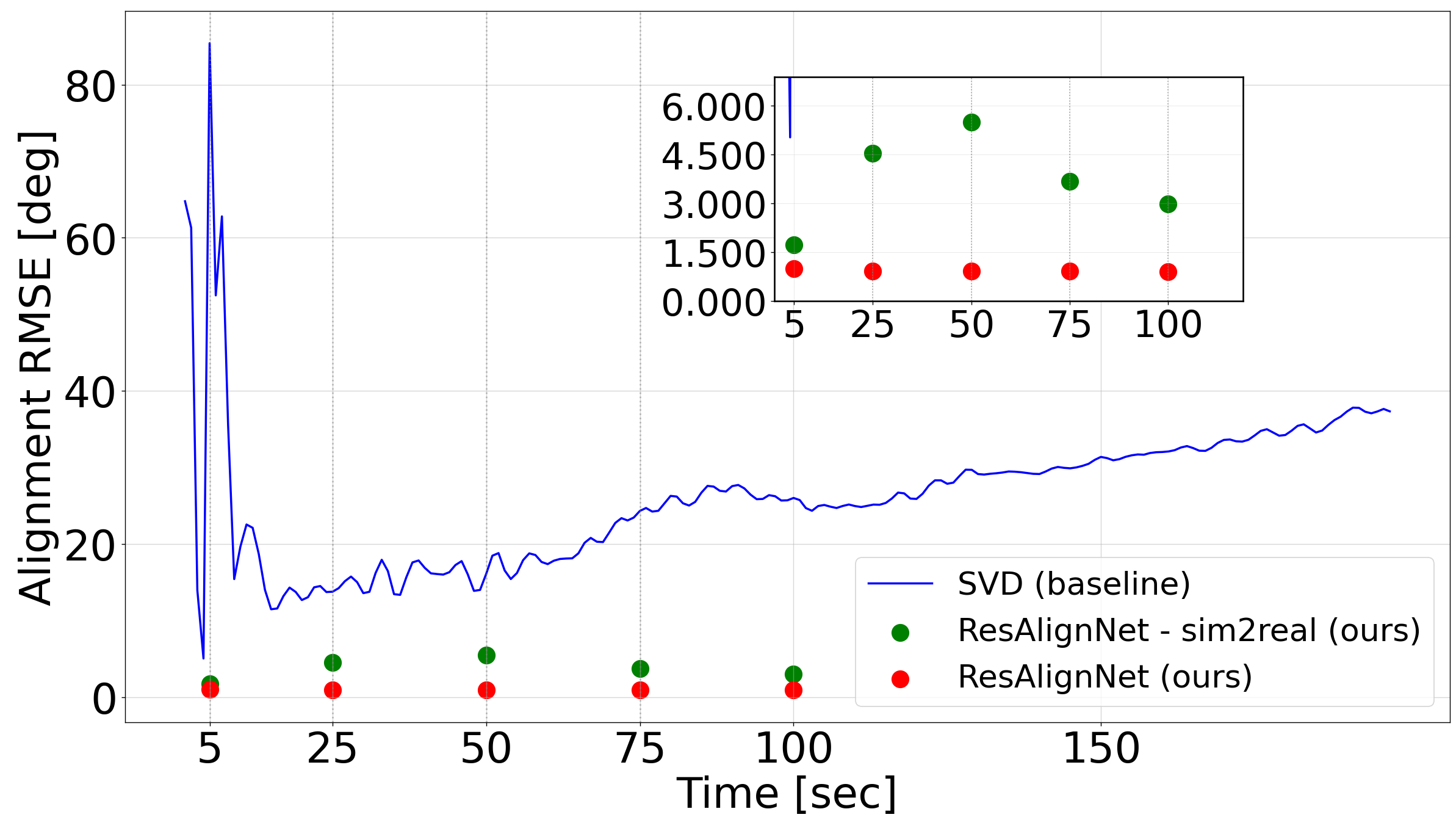}
        \caption{Alignment RMSE performance for Trajectory \#1.}
        \label{fig:real_straight_line_small_angles}
    \end{subfigure}
    
    \vspace{0.5em}
    
    \begin{subfigure}{\columnwidth}
        \centering
        \includegraphics[width=\textwidth]{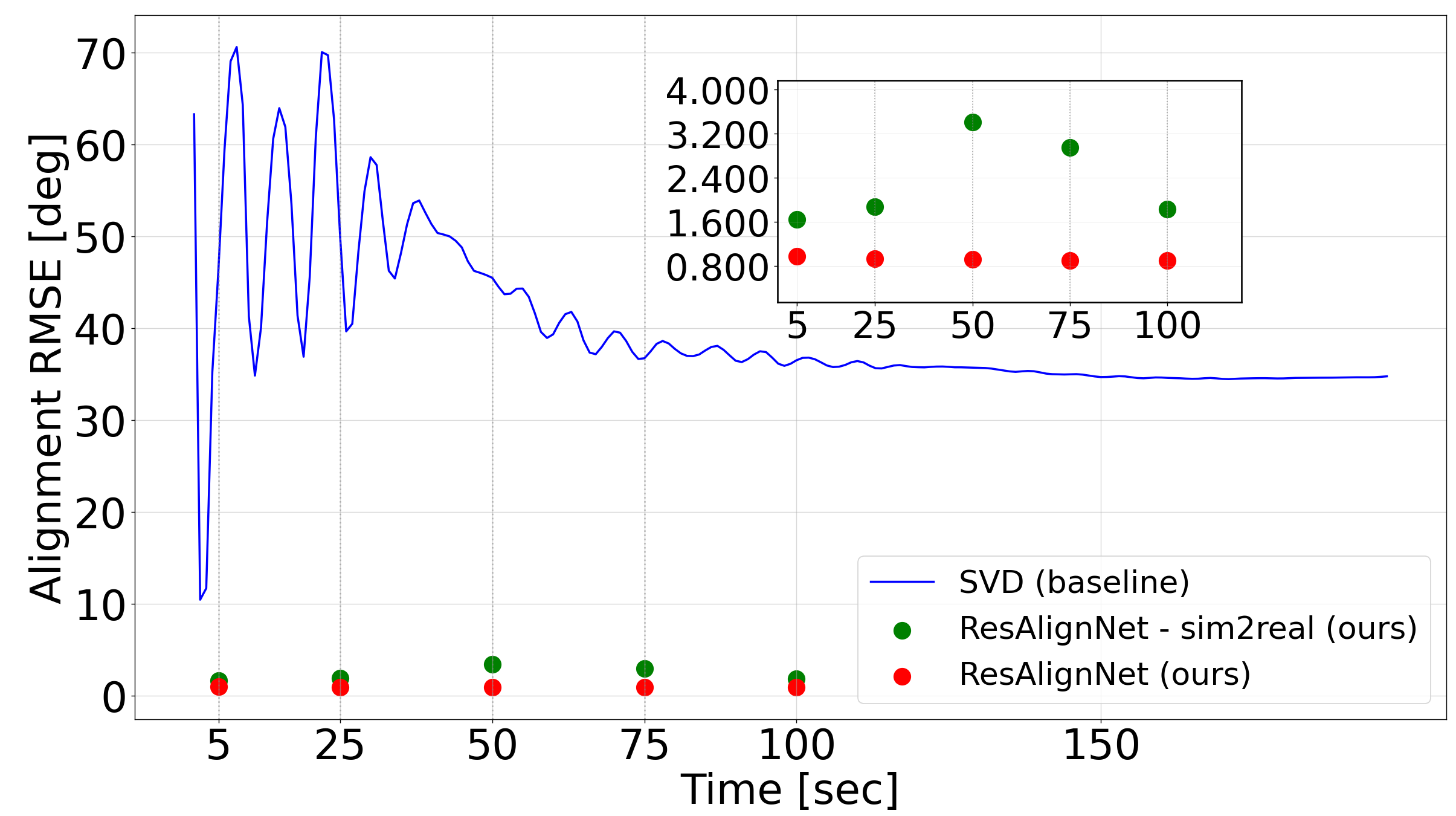}
        \caption{Alignment RMSE performance for Trajectory \#2.}
        \label{fig:real_turn_small_angles}
    \end{subfigure}
    \caption{RMSE alignment performance for an alignment range of 0-5 degrees comparing ResAlignNet and SVD methods across navigation-grade and tactical-grade IMU configurations. The green dots show the results when ResAlignNet was trained on simulated data (Sim2Real), while the red dots represent training on real-world dataset.}
    \label{fig:real_small_misalignment_results}
\end{figure}

\begin{figure}[!h]
    \centering
    \includegraphics[width=\columnwidth]{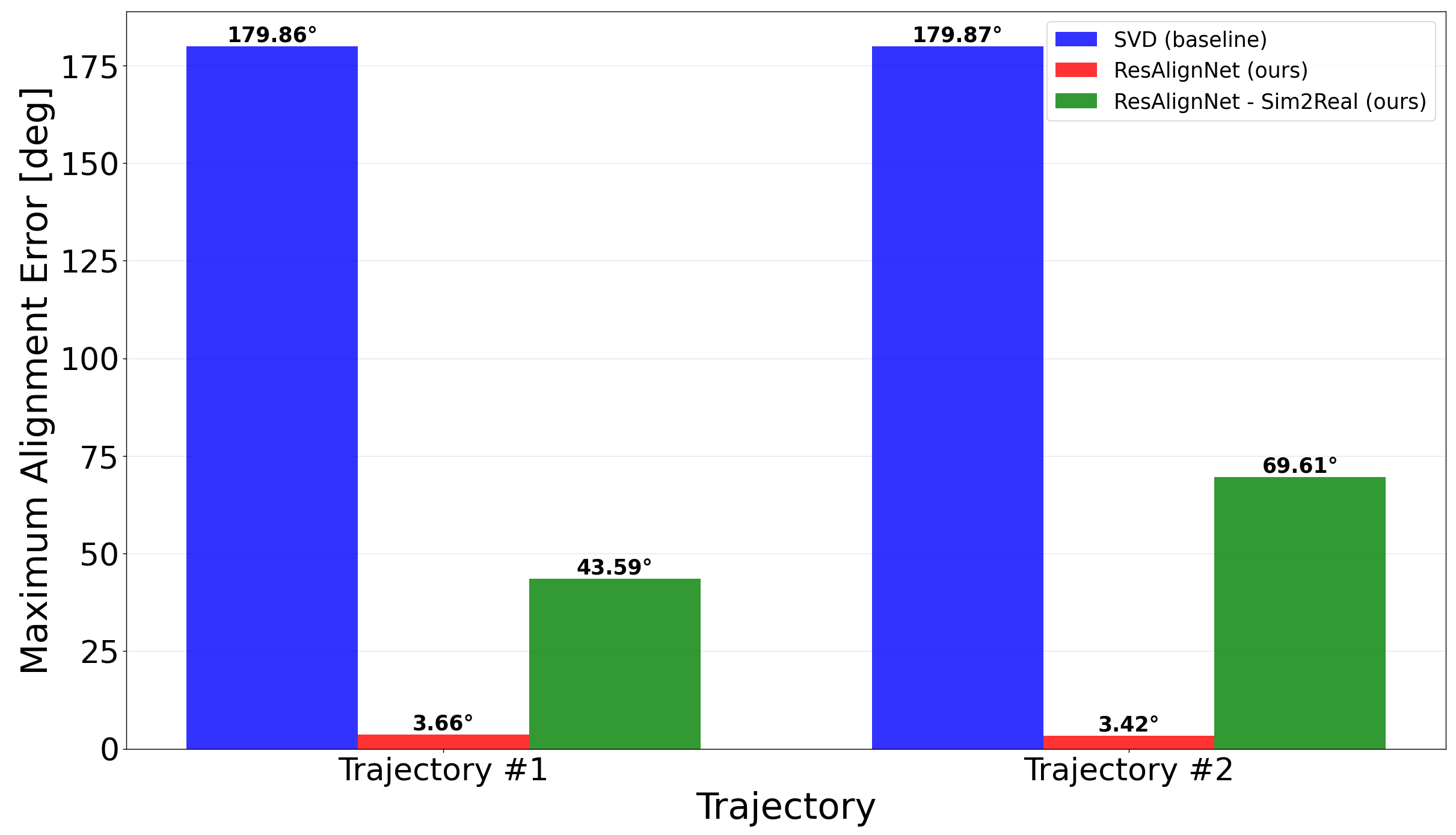}
    \caption{Maximum alignment error comparison between SVD baseline and ResAlignNet methods for Trajectories \#1 and \#2 across the entire test dataset.}
    \label{fig:max_error}
\end{figure}

\noindent
Performance evaluation employs real-world sensor data under operational conditions that reflect tactical-grade IMU specifications used on board the Snapir AUV. This configuration is aligned with the lower-quality sensor parameters examined in the simulation study, allowing direct validation of the predicted performance characteristics.\\
\noindent
Figure \ref{fig:real_small_misalignment_results} presents the experimental results for Trajectory \#1 and Trajectory \#2 using only the real world dataset. ResAlignNet trained on real data achieves exceptional performance with RMSE values below 1 degree for both trajectories across all evaluation windows. 
To evaluate the transferability of learned features from simulation to real-world deployment, we also assess our Sim2Real trained model, a network trained exclusively on simulated data and tested on the identical real-world test dataset. This approach demonstrates effective knowledge transfer, maintaining RMSE values between 1.6-3.4 degrees, although with some performance degradation compared to real data training.\\
\noindent
The model-based SVD performance exhibits significant deterioration and instability consistent with simulation predictions for tactical-grade sensors. Trajectory \#1 demonstrates highly unstable behavior, briefly achieving a minimum RMSE of approximately 13.8 degrees at 20 seconds before rapidly degrading and fluctuating throughout the remaining evaluation period. Trajectory \#2 shows initially erratic performance with oscillations, eventually stabilizing around 35 degrees after 100 seconds, a level that still indicates poor alignment accuracy. This experimental behavior confirms the severe limitations of model-based approaches when deployed with tactical-grade sensors, highlighting both their poor absolute performance and lack of convergence stability under realistic operational conditions.\\
\noindent
Figure \ref{fig:max_error} presents the maximum alignment error comparison across the entire test dataset for both trajectories. ResAlignNet trained on real data maintains maximum errors below 4° (3.66° for Trajectory \#1 and 3.42° for Trajectory \#2), while the Sim2Real approach achieves maximum errors of 43.59° and 69.61° respectively. In contrast, the SVD baseline exhibits severe worst-case performance with maximum errors exceeding 179° for both trajectories, demonstrating the critical reliability advantage of the proposed deep learning approach under challenging alignment scenarios.\\
\noindent
Table \ref{tab:aoe_results} presents the AOE metric for both trajectories in different evaluation windows. The results demonstrate that ResAlignNet consistently achieves sub-degree AOE values, with the real-data trained model maintaining AOE below 0.22° across all window sizes for both trajectories. The Sim2Real trained model shows slightly higher but still competitive performance, with AOE values ranging from 0.39° to 6.02°. In contrast, the baseline SVD exhibits severe performance degradation, with AOE values exceeding 23° for Trajectory \#1 and 63° for Trajectory \#2, confirming the superior alignment accuracy of the proposed deep learning approach.
\begin{table}[!h]
\caption{AOE (degrees) comparison across different window sizes for Trajectories \#1 and \#2. Results show average orientation error for SVD baseline, ResAlignNet Sim2Real, and ResAlignNet trained on real data.}
\label{tab:aoe_results}
\centering
\adjustbox{width=\columnwidth,center}
{
\begin{tabular}{|c|c|c|c|c|}
\hline
\textbf{Trajectory} & \textbf{Window} & \textbf{SVD} & \textbf{ResAlignNet -} & \textbf{ResAlignNet} \\
\textbf{\#} & \textbf{Size (s)} & \textbf{Baseline} & \textbf{Sim2Real (ours)} & \textbf{(ours)} \\
\hline
\multirow{5}{*}{1} & 5 & 147.95 & 0.39 & \textbf{0.23} \\
 & 25 & 23.91 & 3.53 & \textbf{0.21} \\
 & 50 & 28.44 & 3.99 & \textbf{0.21} \\
 & 75 & 43.57 & 3.06 & \textbf{0.21} \\
 & 100 & 46.92 & 3.05 & \textbf{0.21} \\
\hline
\multirow{5}{*}{2} & 5 & 79.45 & 0.41 & \textbf{0.22} \\
 & 25 & 85.32 & 1.56 & \textbf{0.21} \\
 & 50 & 77.76 & 3.93 & \textbf{0.21} \\
 & 75 & 63.55 & 6.02 & \textbf{0.20} \\
 & 100 & 63.90 & 4.33 & \textbf{0.20} \\
\hline
\end{tabular}
}
\end{table}

\noindent

%
\section{Conclusions}\label{conc_sec}
This paper presents ResAlignNet, a data-driven approach to INS/DVL alignment that addresses the convergence time and motion pattern dependency limitations of traditional model-based methods. Using a 1D ResNet-18 architecture optimized for temporal sensor fusion, ResAlignNet operates as an in situ solution requiring only onboard sensors without external positioning aids. The ResNet-18 architecture's deeper hierarchical feature extraction and skip connections prove essential for handling the increased complexity and noise characteristics of real-world sensor data compared to idealized simulation conditions.
\noindent
Experimental validation using the Snapir AUV demonstrates that ResAlignNet achieves alignment accuracy within 0.8° using only 25 seconds of data, representing a 65\% reduction in convergence time compared to velocity-based methods. The approach maintains consistent sub-degree RMSE performance across different sensor quality grades and trajectory patterns, with maximum errors below 4° across the entire test dataset. Traditional SVD methods exhibit severe degradation with tactical-grade sensors, exceeding 35° RMSE under operational conditions with maximum errors exceeding 179°.\\
\noindent
The trajectory-independent nature of ResAlignNet eliminates the need for prescribed motion patterns required by conventional approaches, enabling immediate vehicle deployment without lengthy premission alignment procedures. Sim2Real transfer learning validation demonstrates effective knowledge transfer from simulation to real-world deployment, achieving 1.6-3.4° RMSE when trained exclusively on synthetic data, reducing dependency on extensive real-world data collection.\\
\noindent
ResAlignNet provides a robust sensor-agnostic alignment solution that scales effectively across different operational scenarios and sensor specifications. This research advances underwater navigation capabilities by demonstrating the viability of deep learning for critical sensor fusion tasks, establishing frameworks applicable to broader underwater robotics applications, while significantly reducing operational costs through faster deployment cycles.

\section*{Acknowledgments}
G. D. is grateful for the support of the Maurice Hatter foundation and the University of Haifa excellence scholarship for PhD studies.


\bibliographystyle{IEEEtran}
\bibliography{References.bib}

\end{document}